\documentclass{article} 
\usepackage{iclr2026_conference,times}

\usepackage{amsmath,amsfonts,bm}

\def\eqref#1{equation~\ref{#1}}

\def\1{\bm{1}}

\DeclareMathAlphabet{\mathsfit}{\encodingdefault}{\sfdefault}{m}{sl}
\SetMathAlphabet{\mathsfit}{bold}{\encodingdefault}{\sfdefault}{bx}{n}

\usepackage{hyperref}
\usepackage{url}
\usepackage{graphicx}
\usepackage{booktabs}
\usepackage{caption}

\title{LogitScope: A Framework for Analyzing LLM Uncertainty Through Information Metrics}

\author{Farhan Ahmed \\
  IBM Research \\
  \texttt{farhan.ahmed@ibm.com} \\\And
  Yuya Jeremy Ong \\
  Plastic Labs \\
  \texttt{yuya@plasticlabs.ai} \\\And
  Chad DeLuca \\
  IBM Research \\
  \texttt{delucac@us.ibm.com}
}

\iclrfinalcopy 
\begin{document}

\maketitle


\begin{abstract}

Understanding and quantifying uncertainty in large language model (LLM) outputs is critical for reliable deployment. However, traditional evaluation approaches provide limited insight into model confidence at individual token positions during generation. To address this issue, we introduce \textbf{LogitScope}, a lightweight framework for analyzing LLM uncertainty through token-level information metrics computed from probability distributions. By measuring metrics such as entropy and varentropy at each generation step, LogitScope reveals patterns in model confidence, identifies potential hallucinations, and exposes decision points where models exhibit high uncertainty, all without requiring labeled data or semantic interpretation. We demonstrate LogitScope's utility across diverse applications including uncertainty quantification, model behavior analysis, and production monitoring. The framework is model-agnostic, computationally efficient through lazy evaluation, and compatible with any HuggingFace model, enabling both researchers and practitioners to inspect LLM behavior during inference.

\end{abstract}


\section{Introduction}

Large language models (LLMs) have achieved remarkable capabilities in text generation~\citep{vaswani2017attention,brown2020language,allal2025smollm2}, yet understanding when and why they produce uncertain, incorrect, or unexpected outputs remains challenging~\citep{liang2023holistic,ji2023survey}. Unlike traditional machine learning systems with well-defined prediction tasks and confidence scores, LLMs generate sequences autoregressively where uncertainty manifests across multiple dimensions: individual token probabilities, distribution shapes, and temporal patterns across generation steps. Quantifying this uncertainty is essential for reliable deployment~\citep{zhang2024monitoring,paleyes2022challenges}, debugging model failures, and understanding model knowledge.

Current approaches to uncertainty quantification in LLMs fall into two broad categories. First, aggregate evaluation metrics computed over benchmark datasets~\citep{gao2021framework,liang2023holistic} provide limited insight into token-level behavior during individual inference runs. Second, semantic-level approaches such as self-consistency~\citep{wang2023selfconsistency} or external verification models~\citep{manakul2023selfcheckgpt,min2023factscore} require multiple forward passes or additional models, introducing computational overhead and their own sources of error. Neither approach provides direct, interpretable access to the model's internal uncertainty at each generation step.

We introduce \textbf{LogitScope}, a lightweight framework that addresses these limitations by analyzing token probability distributions during inference to quantify uncertainty through information metrics. At each generation step, language models produce a distribution over their vocabulary. LogitScope computes entropy, varentropy, and surprisal from these distributions to reveal meaningful patterns in model behavior. High entropy indicates broad uncertainty across many tokens, high varentropy suggests multimodal distributions where the model considers distinct alternatives, and high surprisal on selected tokens flags statistically unexpected outputs. These metrics require no labeled data, no additional model calls, and can be computed efficiently in real-time, making them applicable to both research and production settings. We release LogitScope as open-source software to enable the broader community to analyze LLM uncertainty in their applications\footnote{Code available at: \url{https://github.com/ibm-granite/granite.debug-tools/tree/main/logitscope}}.

Our contributions are as follows:
\begin{itemize}
    \item We present LogitScope, an open-source framework that quantifies token-level LLM uncertainty by computing information metrics such as entropy and varentropy from probability distributions. This requires no labeled data, additional models, or multiple forward passes.
    \item We demonstrate how these information metrics provide interpretable signals about model confidence, revealing patterns associated with hallucinations, decision points, and unexpected outputs across diverse applications including model analysis, debugging, and production monitoring.
\end{itemize}

\section{Background and Related Work}

\paragraph{Uncertainty Quantification in LLMs} Recent work has explored various approaches to quantifying uncertainty in language models. Perplexity and token probability have long been used as confidence measures~\citep{jelinek1977perplexity,brown2020language}, but provide limited insight into distribution characteristics. Self-consistency methods~\citep{wang2023selfconsistency} generate multiple outputs and measure agreement, but require multiple forward passes. Semantic uncertainty approaches~\citep{kuhn2023semantic} cluster model outputs in semantic space, but depend on external models and are computationally expensive. Entropy-based metrics have been explored in specific contexts~\citep{malinin2018predictive,lin2023generating}, but typically focus on individual metrics in isolation for specific tasks. Hallucination detection methods~\citep{manakul2023selfcheckgpt,ji2023survey,min2023factscore} similarly rely on multiple generations or external knowledge bases. LogitScope provides a unified framework for computing and analyzing multiple information metrics across diverse uncertainty quantification and model analysis applications.

\paragraph{Model Monitoring and Drift Detection} The ML monitoring community has developed extensive tooling for supervised learning systems~\citep{paleyes2022challenges,zhang2024monitoring}, focusing on input drift, prediction drift, and performance degradation. However, these approaches assume access to labels and well-defined prediction tasks. For generative language models, defining appropriate monitoring signals is more challenging due to open-ended outputs and subjective quality assessment~\citep{liang2023holistic,gao2021framework}. Recent work has explored self-supervised signals for LLM monitoring, including consistency checks~\citep{manakul2023selfcheckgpt,wang2023selfconsistency}, factuality scoring~\citep{min2023factscore}, and attention pattern analysis~\citep{darcet2024vision}. These methods often require multiple forward passes~\citep{wang2023selfconsistency} or external verification systems~\citep{min2023factscore,manakul2023selfcheckgpt}. LogitScope complements these approaches by providing lightweight, interpretable metrics that require no additional model calls or external knowledge bases.

\section{Method}

\subsection{Information Metrics for Uncertainty Analysis}

LogitScope computes metrics from the probability distribution $p_t = p(x_t | x_{<t})$ at each token position $t$ (formal definitions in Appendix~\ref{sec:metric_definitions}):

\begin{itemize}
    \item \textbf{Probability}: Direct confidence measure; the model's assigned probability for the selected token.
    \item \textbf{Surprisal}: Negative log-probability of the selected token; quantifies how unexpected the choice was given the context.
    \item \textbf{Entropy}: Sum of weighted surprisals across all tokens; measures overall uncertainty in the distribution.
    \item \textbf{Varentropy}: Variance of surprisal values; high varentropy with high entropy indicates multimodal distributions where the model considers distinct alternatives.
    \item \textbf{Skewentropy}: Distribution asymmetry; reveals whether probability mass is concentrated or dispersed.
    \item \textbf{Perplexity}: Exponential of average surprisal; provides cumulative sequence-level quality measure.
\end{itemize}

\subsection{Implementation}

LogitScope is implemented as a lightweight wrapper around HuggingFace Transformers~\citep{wolf2020transformers}. The core \texttt{LogitScope} class takes a tokenizer and model, performs inference on input text, and returns a \texttt{Results} object containing the probability distributions and lazy-evaluated metrics.

The framework uses two key design principles:
\begin{enumerate}
    \item \textbf{Lazy evaluation}: Metrics are computed on-demand and cached, avoiding unnecessary computation when only specific metrics are needed.
    \item \textbf{Zero-copy access}: Raw logits and probability distributions are accessible for custom analysis without copying data.
\end{enumerate}

The framework supports CPU, CUDA, and Apple Silicon (MPS) acceleration and works with any HuggingFace model without modification.

\section{Analysis and Applications}

\subsection{Uncertainty Pattern Analysis}

\begin{figure}[t]
    \centering
    \begin{minipage}[t]{0.52\columnwidth}
        \vspace{0pt}
        \centering
        \includegraphics[width=\linewidth]{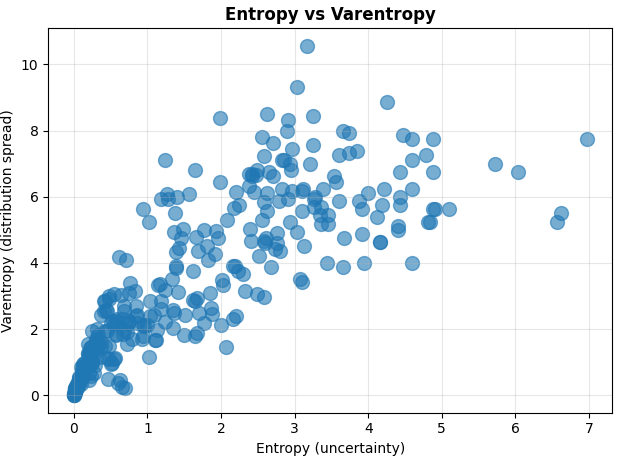}
        \caption{Entropy vs varentropy scatter plot. Each point represents a token position, with lower-left indicating confident predictions and upper-right indicating multimodal uncertainty.}
        \label{fig:entropy_varentropy}
    \end{minipage}
    \hfill
    \begin{minipage}[t]{0.45\columnwidth}
        \vspace{0pt}
        \centering
        \small
        \captionof{table}{Average metrics for the Declaration of Independence preamble with original vs word-reversed order. Reversing the word order destroys coherence, dramatically increasing uncertainty.}
        \label{tab:metrics_comparison}
        \vspace{0.3em}
        \begin{tabular}{lcc}
        \toprule
        \textbf{Metric} & \textbf{Original} & \textbf{Reversed} \\
        \midrule
        Tokens & 320 & 320 \\
        Characters & 1628 & 1628 \\
        \midrule
        Entropy & 1.70 & 6.33 \\
        Varentropy & 3.46 & 8.63 \\
        Skewentropy & 7.50 & 0.48 \\
        Perplexity & 11.75 & 1833.76 \\
        Probability & 0.55 & 0.09 \\
        Log Probability & -1.05 & -2.70 \\
        \bottomrule
        \end{tabular}
    \end{minipage}
\end{figure}

We demonstrate LogitScope's analytical capabilities through two complementary views of the Declaration of Independence preamble (320 tokens; see Appendix~\ref{sec:eval_text}) processed by SmolLM2-135M-Instruct~\citep{allal2025smollm2}, a compact 135M parameter model. Figure~\ref{fig:entropy_varentropy} reveals token-level uncertainty patterns through entropy and varentropy, while Table~\ref{tab:metrics_comparison} quantifies how destroying linguistic structure by reversing word order affects aggregate model confidence.

\paragraph{Token-level patterns} Figure~\ref{fig:entropy_varentropy} shows that tokens cluster into distinct regions. \textit{Low entropy and low varentropy} (bottom-left) represents confident predictions where the model assigns high probability to a single token. Common words and grammatically constrained positions fall into this region, indicating strong prior expectations. \textit{High entropy and high varentropy} (top-right) indicates multimodal uncertainty, where the model distributes probability mass across multiple plausible alternatives. This pattern often appears at semantic decision points when multiple valid continuations exist. \textit{Intermediate regions} reveal gradations of confidence between these extremes.

\paragraph{Aggregate patterns} Table~\ref{tab:metrics_comparison} compares the original and word-reversed text which reveals how linguistic structure affects model confidence. Reversing word order destroys both semantic coherence and syntactic dependencies, forcing the model to process grammatically invalid sequences. The reversed text shows significantly higher values for entropy (6.33 vs. 1.70), varentropy (8.63 vs 3.46), perplexity (1833.76 vs. 11.75), while average token probability drops from 0.55 to 0.09. This demonstrates LogitScope's ability to quantify how language structure influences model uncertainty across entire sequences.

\subsection{Applications}

Beyond understanding uncertainty patterns, LogitScope's information metrics support diverse practical workflows:

\textbf{Hallucination detection:} High entropy and varentropy regions often correlate with hallucinated content, as models exhibit uncertainty when generating facts beyond their training data. By flagging tokens with unusual metric patterns, practitioners can identify outputs requiring verification.

\textbf{Model debugging:} When models produce unexpected outputs, LogitScope reveals whether the issue stems from low-confidence predictions (high entropy), competition between alternatives (high varentropy), or statistically unlikely selections (high surprisal). This diagnostic information guides debugging efforts.

\textbf{Prompt engineering:} Comparing metric distributions across different prompt formulations reveals which prompts elicit more confident predictions on correct outputs. Effective prompts typically reduce entropy while maintaining high probability on expected tokens.

\textbf{Production monitoring:} Aggregate statistics (mean entropy, median surprisal, etc.) provide real-time signals about model behavior. Sudden shifts in these distributions can indicate input drift, adversarial inputs, or edge cases requiring attention, all without accessing ground truth labels.

\textbf{Model comparison:} LogitScope enables quantitative comparison of different models on the same inputs. Beyond accuracy metrics, practitioners can assess whether models differ in their uncertainty patterns, confidence calibration, or decision-making strategies.

\section{Limitations and Future Work}

\textbf{Limitations:} Information metrics provide signals about distribution characteristics and model uncertainty but do not directly measure semantic correctness or factual accuracy. High confidence (low entropy) does not guarantee correct outputs; models can be confidently wrong. Similarly, high entropy does not always indicate errors; it may reflect genuine ambiguity. LogitScope is best viewed as an analysis tool that reveals uncertainty patterns for human inspection, rather than an automated correctness verifier. The framework currently focuses on token-level analysis and does not capture longer-range semantic patterns or factual consistency.

\textbf{Broader Impact:} By providing interpretable signals about model uncertainty, LogitScope can help practitioners identify when models are operating outside their reliable range, potentially reducing the deployment of overconfident but incorrect outputs. However, the metrics can also be gamed through careful prompt engineering to artificially reduce entropy without improving actual correctness.

\textbf{Future Directions:} The framework can be extended with additional metrics tailored to specific failure modes, integrated into MLOps pipelines for automated anomaly detection, or combined with semantic clustering approaches for comprehensive uncertainty quantification. Large-scale empirical studies correlating metric patterns with human quality judgments across diverse domains and model architectures would further validate the utility of information uncertainty measures.

\section{Conclusion}

We presented LogitScope, a lightweight framework for analyzing LLM uncertainty through information metrics computed from token probability distributions. By measuring entropy, varentropy, and surprisal at each generation step, LogitScope enables researchers and practitioners to quantify model confidence, identify uncertainty patterns, and detect potential issues in real-time without requiring labeled data or additional model calls. We demonstrated the framework's utility across diverse applications and released it as open-source software to support the community in understanding and improving LLM reliability.

\bibliography{references}
\bibliographystyle{iclr2026_conference}

\newpage

\appendix

\section{Metric Definitions}
\label{sec:metric_definitions}

At each generation step $t$, the model produces logits $\mathbf{z}_t \in \mathbb{R}^{|V|}$ over vocabulary $V$, which are transformed into a probability distribution via softmax:
\begin{equation}
p_t(v) = p(x_t = v | x_{<t}) = \frac{\exp(z_t^{(v)})}{\sum_{v' \in V} \exp(z_t^{(v')})}
\end{equation}

LogitScope computes the following metrics from this probability distribution $p_t$ at each token position $t$:

\paragraph{Probability} is the direct confidence measure for the selected token:
\begin{equation}
P(x_t) = p_t(x_t)
\end{equation}
High probability indicates confident selection, while low probability suggests the token was unlikely given the context.

\paragraph{Surprisal} quantifies how unexpected the selected token $x_t$ was:
\begin{equation}
I(x_t) = -\log p_t(x_t)
\end{equation}
Surprisal is the fundamental building block for other metrics. High surprisal indicates the model assigned low probability to the token it ultimately selected, signaling statistically unlikely outputs.

\paragraph{Entropy} measures the overall uncertainty in the distribution:
\begin{equation}
H(p_t) = -\sum_{v \in V} p_t(v) \log p_t(v) = \sum_{v \in V} p_t(v) \cdot I(v)
\end{equation}
Entropy is the expected surprisal across all tokens. High entropy indicates broad uncertainty across many possible tokens, while low entropy suggests confident predictions concentrated on few tokens.

\paragraph{Varentropy} measures the variance of surprisal values across the distribution:
\begin{equation}
\text{Var}(p_t) = \sum_{v \in V} p_t(v) \left(\log p_t(v)\right)^2 - H(p_t)^2
\end{equation}
High varentropy combined with high entropy indicates a multimodal distribution where the model is torn between multiple distinct options, which often occurs at decision points or when hallucinating.

\paragraph{Skewentropy} measures the asymmetry of the surprisal distribution:
\begin{equation}
\text{Skew}(p_t) = \frac{\sum_{v \in V} p_t(v) \left(\log p_t(v) + H(p_t)\right)^3}{\text{Var}(p_t)^{3/2}}
\end{equation}
Skewentropy reveals whether probability mass is concentrated (high absolute skew) or uniformly distributed (low skew), providing insight into distribution shape beyond entropy and varentropy.

\paragraph{Perplexity} provides a cumulative measure of model performance over the sequence:
\begin{equation}
\text{PPL} = \exp\left(\frac{1}{T}\sum_{t=1}^{T} I(x_t)\right)
\end{equation}
Perplexity exponentiates the average surprisal, yielding an interpretable measure of predictive quality. Lower perplexity indicates better model performance.

\section{Evaluation Text}
\label{sec:eval_text}

All examples in this paper use the preamble of the United States Declaration of Independence as the evaluation text, processed by SmolLM2-135M-Instruct~\citep{allal2025smollm2}. The complete text (320 tokens when tokenized by SmolLM2) is:

\begin{quote}
\small
We hold these truths to be self-evident, that all men are created equal, that they are endowed by their Creator with certain unalienable Rights, that among these are Life, Liberty and the pursuit of Happiness.--That to secure these rights, Governments are instituted among Men, deriving their just powers from the consent of the governed, --That whenever any Form of Government becomes destructive of these ends, it is the Right of the People to alter or to abolish it, and to institute new Government, laying its foundation on such principles and organizing its powers in such form, as to them shall seem most likely to effect their Safety and Happiness. Prudence, indeed, will dictate that Governments long established should not be changed for light and transient causes; and accordingly all experience hath shewn, that mankind are more disposed to suffer, while evils are sufferable, than to right themselves by abolishing the forms to which they are accustomed. But when a long train of abuses and usurpations, pursuing invariably the same Object evinces a design to reduce them under absolute Despotism, it is their right, it is their duty, to throw off such Government, and to provide new Guards for their future security.--Such has been the patient sufferance of these Colonies; and such is now the necessity which constrains them to alter their former Systems of Government. The history of the present King of Great Britain is a history of repeated injuries and usurpations, all having in direct object the establishment of an absolute Tyranny over these States. To prove this, let Facts be submitted to a candid world.
\end{quote}

This text was chosen for its historical significance, formal register, complex syntactic structure, and moderate length suitable for demonstration purposes. The word-reversed version used in Table~\ref{tab:metrics_comparison} reverses the order of words while preserving individual word spellings, creating grammatically invalid but tokenizable sequences.

\section{Interactive Web Interface}

In addition to the Python library, LogitScope provides an interactive web-based interface for visual exploration of model uncertainty. The UI enables real-time analysis during text generation, displaying token-level metrics with color-coded visualizations and interactive controls. Users can switch between different metrics, inspect top-k alternative tokens at each position, and observe temporal patterns across the generation sequence.

\begin{figure}[htbp]
    \centering
    \includegraphics[width=0.98\columnwidth]{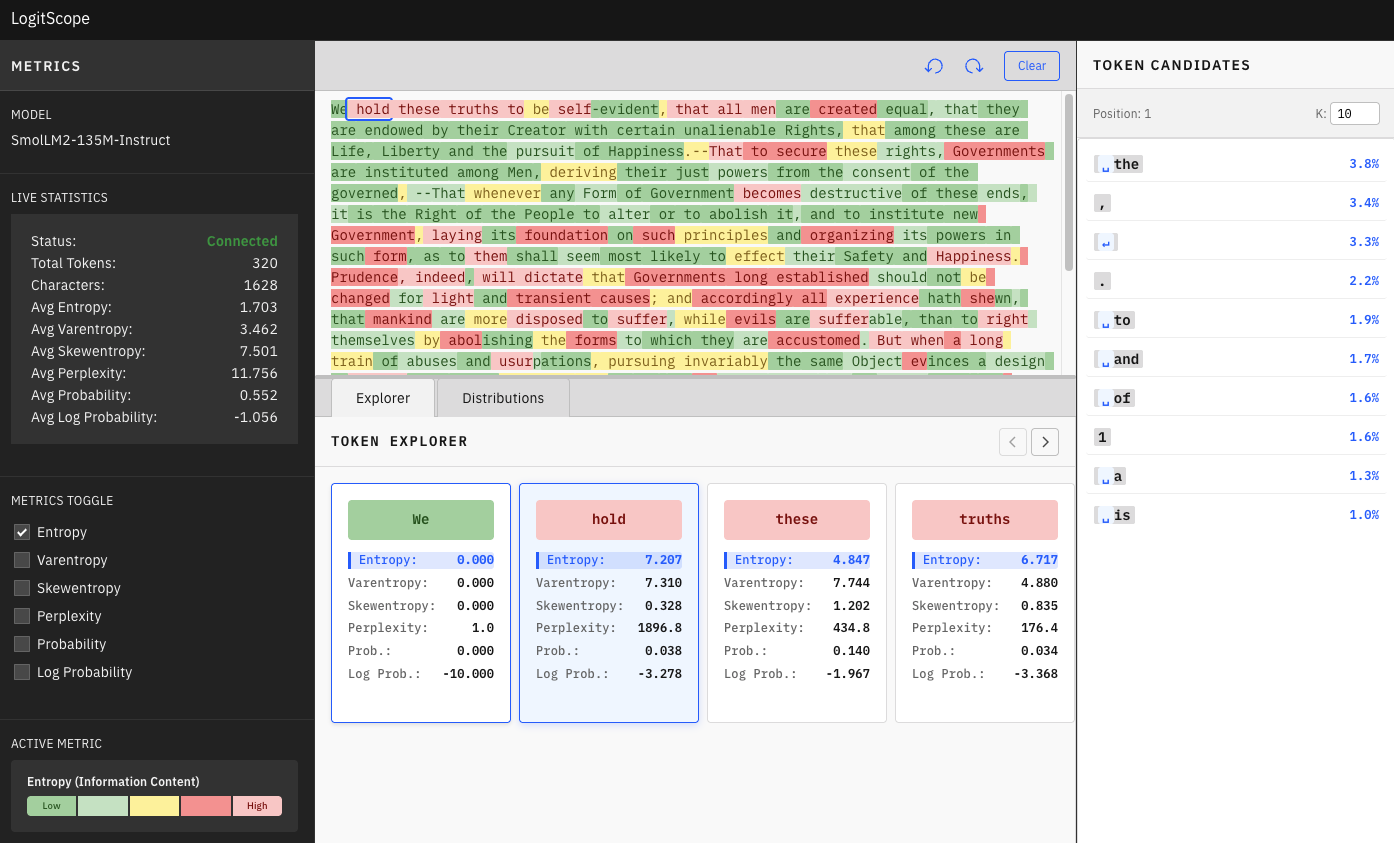}
    \caption{The web interface displaying the entropy metric. Tokens are color-coded by magnitude, with brighter colors indicating higher uncertainty. The sidebar shows aggregate statistics for quick assessment of overall model confidence.}
    \label{fig:ui_entropy}
\end{figure}

\begin{figure}[htbp]
    \centering
    \includegraphics[width=0.98\columnwidth]{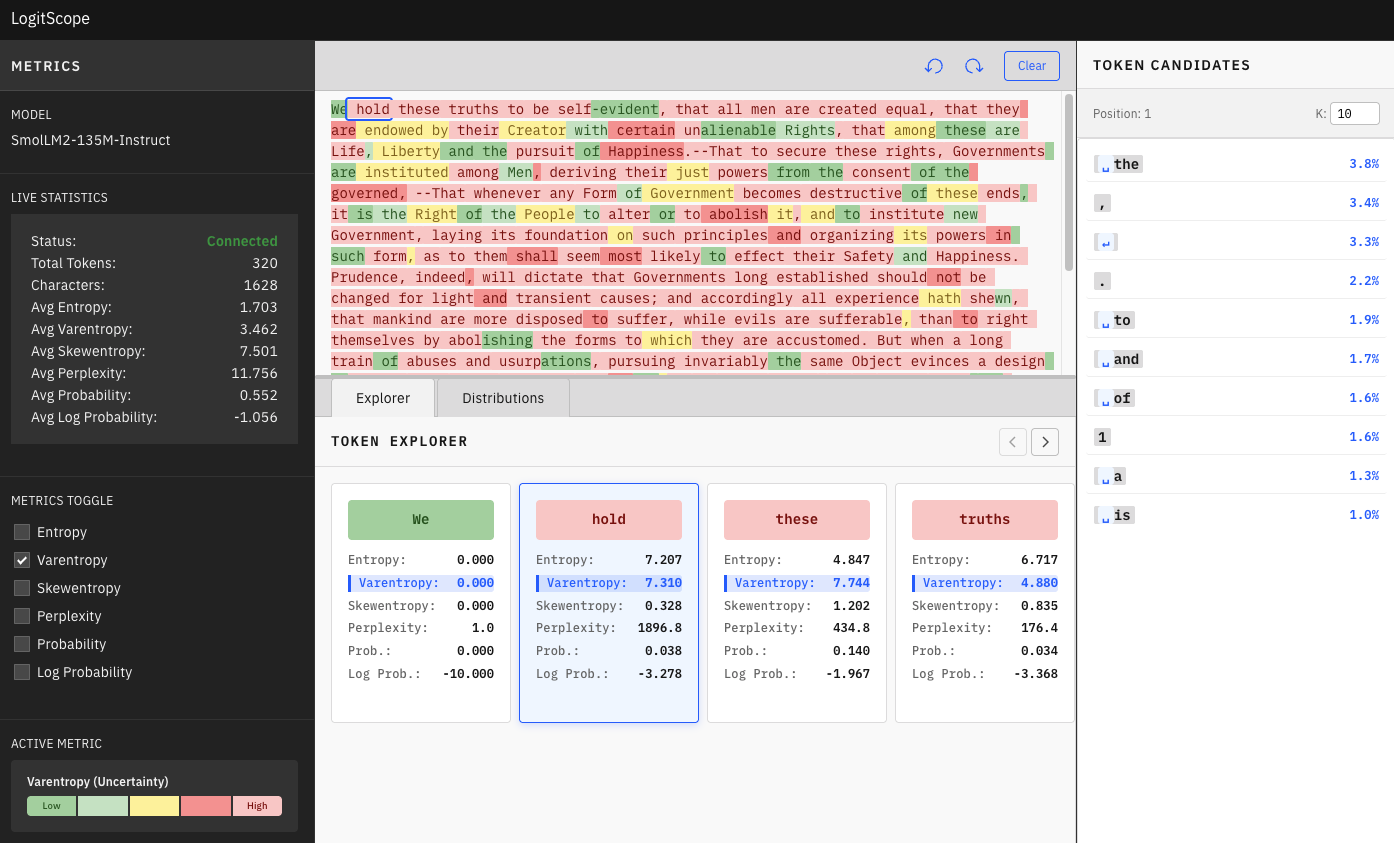}
    \caption{The web interface displaying the varentropy metric. Varentropy reveals multimodal distributions where the model considers multiple alternatives. High varentropy regions indicate decision points with competing continuations.}
    \label{fig:ui_varentropy}
\end{figure}

\begin{figure}[htbp]
    \centering
    \includegraphics[width=0.98\columnwidth]{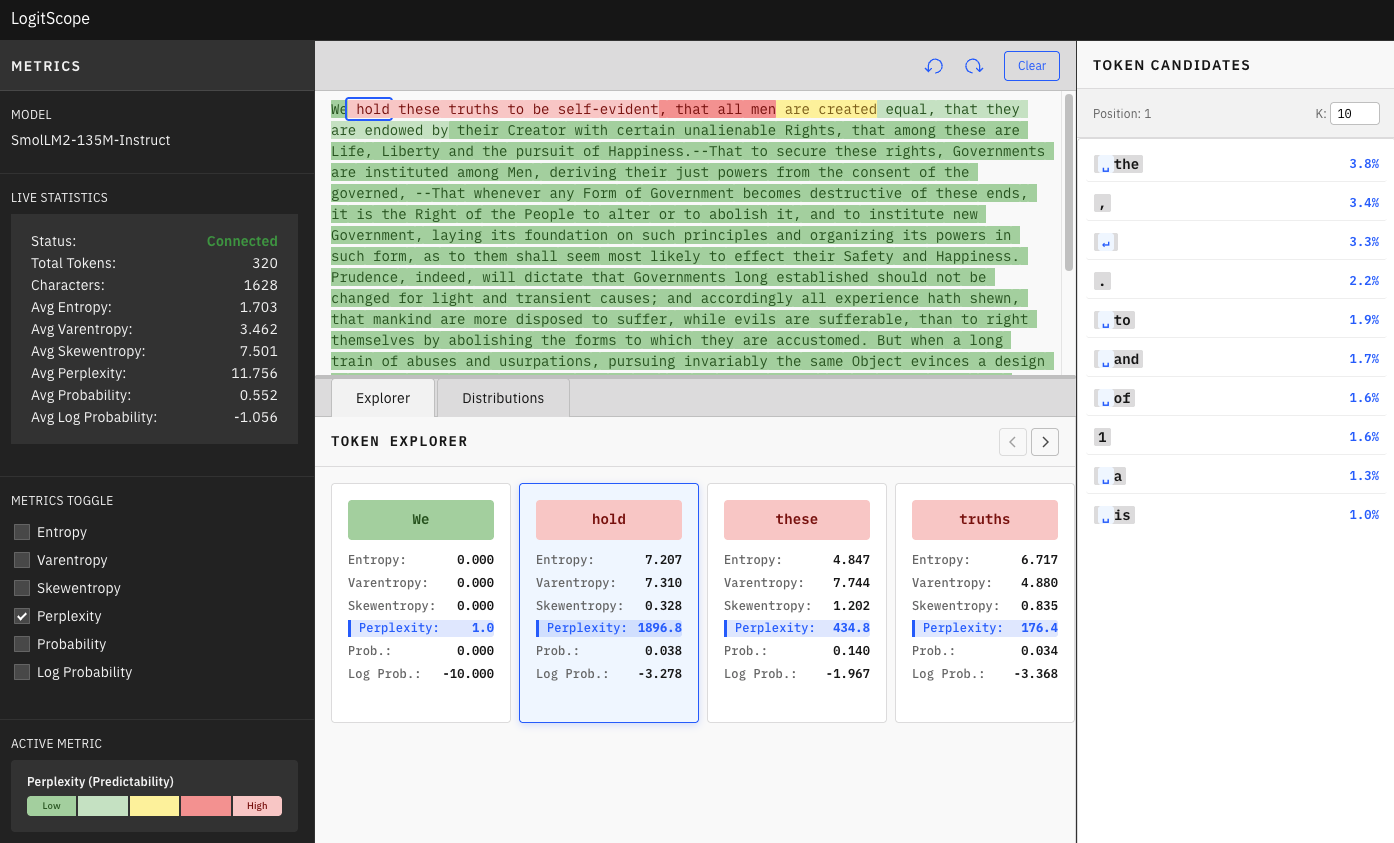}
    \caption{The web interface displaying the perplexity metric. Lower values indicate better predictive confidence. The interface enables identification of regions where the model exhibits uncertainty.}
    \label{fig:ui_perplexity}
\end{figure}

\begin{figure}[htbp]
    \centering
    \includegraphics[width=0.98\columnwidth]{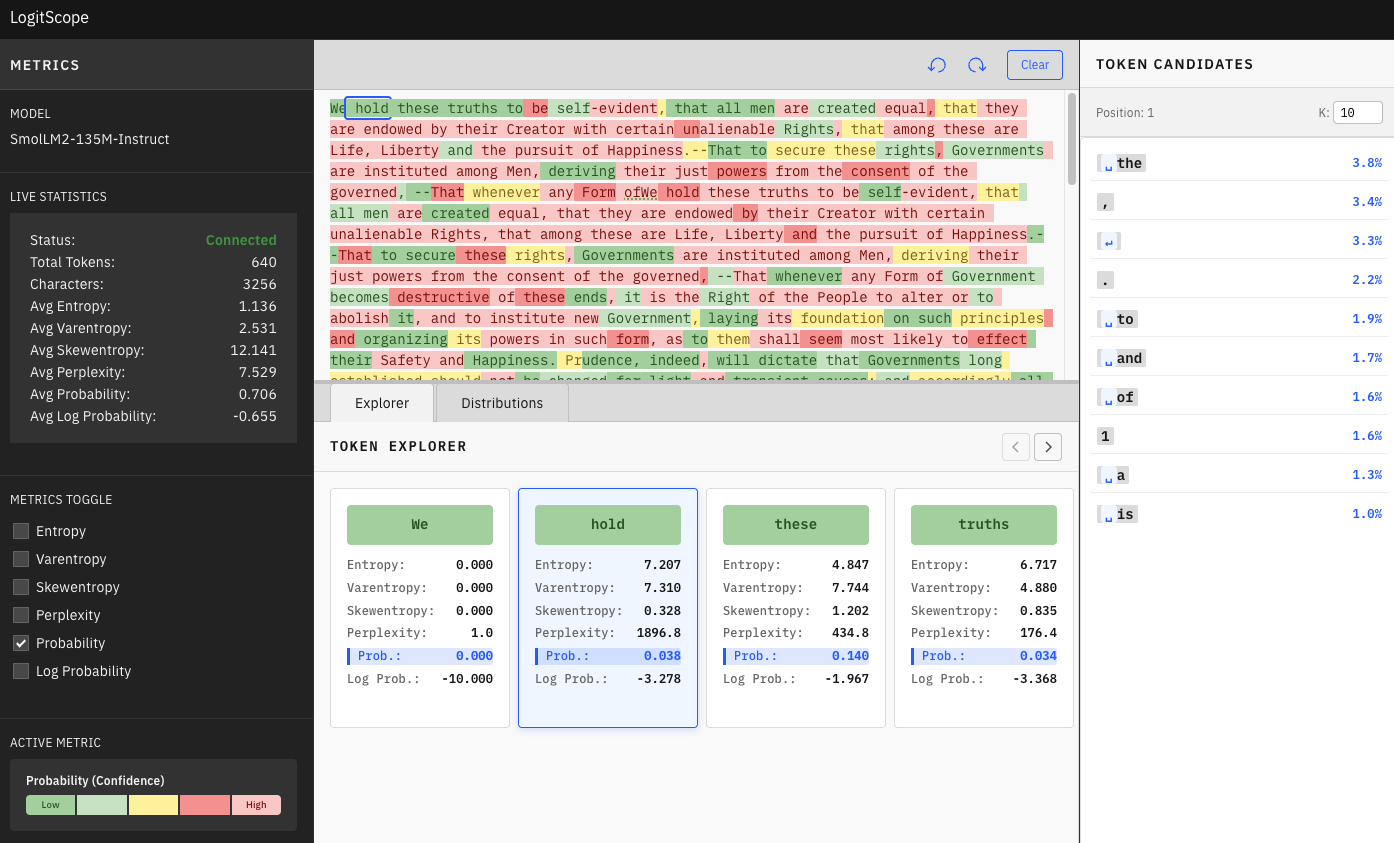}
    \caption{The web interface displaying the probability metric. Higher values indicate more confident token selections. The direct probability measure provides intuitive assessment of model certainty.}
    \label{fig:ui_probability}
\end{figure}

Figures~\ref{fig:ui_entropy}, \ref{fig:ui_varentropy}, \ref{fig:ui_perplexity}, and \ref{fig:ui_probability} show the interface analyzing the Declaration of Independence preamble using SmolLM2-135M-Instruct~\citep{allal2025smollm2}. The interface highlights tokens by their metric values, with color intensity indicating magnitude. The sidebar displays running statistics (mean, median, min, max) and allows users to toggle between different metrics. Clicking on individual tokens reveals the top-k alternatives the model considered at that position, along with their probabilities.

The web interface is particularly useful for:
\begin{itemize}
    \item \textbf{Rapid prototyping}: Quickly test different prompts and observe their effect on model uncertainty without writing code.
    \item \textbf{Educational purposes}: Demonstrate model behavior to students or stakeholders through intuitive visual feedback.
    \item \textbf{Debugging}: Identify specific tokens or regions where models exhibit unexpected uncertainty patterns.
    \item \textbf{Comparative analysis}: Switch between metrics to understand different aspects of the same generation.
\end{itemize}

The interface is a component of the LogitScope framework and is launched directly from it. All models compatible with HuggingFace Transformers are supported with automatic device detection for CPU, CUDA, and Apple Silicon acceleration.

\end{document}